# CEC-Zero: Chinese Error Correction Solution Based on LLM and Reinforcement Learning

Sophie Zhang, Zhiming Lin

sophiezhangshsid@163.com,nklinzhiming@gmail.com

I. Summary:

Recently, the development and advancement of large-scale language models (LLMs) has amazed the entire AI community. Thanks to their emerging capabilities, LLMs have attracted more and more researchers to study their capabilities and performance on various downstream natural language processing (NLP) tasks. While marveling at the amazing performance of LLMs on a variety of tasks, we have noticed that they also have excellent language processing capabilities, such as Chinese. To explore the Chinese processing capabilities of LLMs, we focus on Chinese text correction, a fundamental and challenging task in Chinese NLP. Specifically, we evaluate a variety of representative LLMs on the Chinese Spelling Checking (CSC) task.Currently, LLMs have both impressive performance in Chinese text correction, and the combined performance and robustness of LLMs are much higher than that of the domain-trained miniatures of BERTs, but there are also unsatisfactory areas.

With the widespread use of large language models (LLMs) in text generation tasks, Chinese Spelling correction (CSC) has become a key challenge to improve model reliability. Traditional approaches mainly rely on supervised fine-tuning (SFT) **[18][32]** or multi-model collaboration (e.g., validation models). However, these methods face challenges such as biased data distribution, huge computational effort, and insufficient generalization capability. In this paper, we propose a new framework, called CEC-Zero, which performs self-correction through reinforcement learning, which is based on the combination of Large Language Modeling (LLM) and Reinforcement Learning (RL). The framework aims to facilitate the learning of error correction strategies by the model through self-generated data without the need for external supervision or the use of other models. This research provides an efficient and scalable solution for self-correction of LLMs and a new paradigm for reliability optimization in multi-step inference tasks.

Therefore, we propose here to use reinforcement learning algorithms for secondary training of open-source LLMs to better observe the potential capabilities of LLMs. Through extensive analysis and comparison with previous state-of-the-art LLMs and BERT-type mini-models, we empirically found that RL is able to improve the accuracy and cross-domain generalization of LLMs to an accuracy acceptable and usable by the industry, and we believe that our discovery will facilitate the landing and application of LLMs in the Chinese NLP community.

II. Introduction:

2.1, Textual corrections:

Large-scale language models (LLMs) have made significant progress in the past few months and are gradually becoming the infrastructure for the field of natural language processing (NLP) **[54]**. Benefiting from emerging reasoning capabilities **[23][51][58][60]** and the advantages of the chain of thought, LLMs seem to be sweeping a variety of downstream tasks **[14][24][37][40][48]** in NLP with a unified conversational paradigm. Over the past few months, LLMs have been evaluated extensively on a variety of NLP tasks and have shown performance beyond expectations, such as natural language understanding, information extraction, classification judgments, and text summarization. In addition to LLM's generalized capabilities for various tasks, its excellent multilingual adaptability is also impressive **[5]**. Chinese, as one of the most spoken languages in the world, has always been a challenging and research-worthy language in the NLP community due to its unique linguistic nature and characteristics. Therefore, in this paper, we comprehensively evaluate the ability of LLMs to perform a basic and challenging Chinese NLP task: Chinese text correction.

Chinese Text Error Correction (CEC) is designed to detect and correct various errors contained in the input Chinese text. According to the types of errors, Chinese text error correction is generally categorized into two types of tasks: Chinese Grammar Error Correction (CGEC) and Chinese Spelling Check (CSC). Grammar correction (CGEC) tasks mainly target learners of Chinese as a second language (CSL) who are prone to the language learning process, and native speakers of Chinese who accidentally make grammatical errors in their daily lives. Due to the differences in language levels, the focus of CGEC tasks for CSL learners and CGEC tasks for native Chinese speakers differ. Since CSL learners do not have a high level of Chinese language mastery, the grammatical errors they often make mainly involve the addition, deletion, substitution, and rearrangement of Chinese characters, whereas the grammatical errors made by native Chinese speakers tend to be more subtle and difficult, which poses higher requirements and challenges for the model to understand the Chinese grammar rules, and such scenarios are usually found in order to subjectively circumvent some complex language quality control audit scenarios. As for the CSC task, it is to automatically check spelling errors in Chinese text. Due to the characteristics of Chinese characters, Chinese spelling errors are mainly caused by phonetically or visually similar characters. Therefore, CSC is challenging because it requires not only complex semantic knowledge, but also phonetic/visual information **[7][64][68][72]** to help the model in finding the correct characters. From the above, it can be seen that Chinese character correction is a practical and complex scenario for Chinese applications.

In this paper, we conducted a comprehensive study to evaluate the Chinese error correction ability of LLMs. First, based on the characteristics of CGEC and CSC, we crafted task-specific prompts to guide LLMs to correct errors like an error corrector. Then, we explored some widely used contextual learning prompting strategies to further motivate language teachers to correct errors.

In addition, we construct targeted training data to post-train LLM based on the inadequacy of existing Chinese text error correction training data and the diversity of complexity of the usage scenarios, so as to further improve its error correction capability. We conduct extensive experiments on CSL and native CGEC benchmarks as well as CSC datasets**[52][65]** from multiple domains. It is worth noting that the appeal of textual error correction is that it is a highly subjective task, i.e., there may be multiple reference sentences for an erroneous sentence, and thus automatic evaluations of benchmarks and objective metrics may not truly reflect the performance of the model, and thus we also conducted in-depth manual evaluations to observe the more realistic error-correction capabilities of the LLM.

2.2 LLM:

Test-Time Scaling (TTS) is an emerging trend in enhancing the reasoning capabilities of large language models (LLMs). Recent studies [6][20][26][43][53] have shown that TTS is more computationally efficient than scaling [23] during pre-training and can achieve better performance with the same computational investment. Many studies have explored strategies such as augmenting TTS with reward models, using majority voting and Monte Carlo tree search in the decoding phase. Recent leading Large Reasoning Models (LRMs), such as DeepSeek-R1 [9] and OpenAI's o1[31][39][73] , emphasize the key role of Reinforcement Learning (RL) [38][44] in augmenting long chains of thought. However, LRMs still struggle to address new unlabeled data streams. For example, while OpenAI o3 achieved a 75.7% success rate on ARC-AGI-1, it only solved 4% of the problems on ARC-AGI-2 (2025). These studies on TTS clearly show the difference between training time and testing time behavior, especially in RL-based approaches [8][11][13][25][27][28][36][42][59][66] that focus on training time. However, applying RL only on large-scale training data is grossly insufficient to handle new features or distributional changes in the emerging highly complex inputs.

In recent years, Test-Time Training (TTT) methods [62][56][19] significantly improve the model generalization ability by dynamically updating the model parameters using test data, among which Test-Time Reinforcement Learning (TTRL) [55] combined with Reinforcement Learning (RL) has become an important development direction. However, the method faces a core challenge: it is difficult to obtain reliable reward signals in test scenarios without real labels. As the task complexity increases and the data volume proliferates, it is no longer practical to rely on manual labeling to construct the reward mechanism, which severely restricts the continuous learning capability of the model. For this reason, TTRL innovatively proposes a repeated sampling strategy to generate rule-based pseudo-reward signals through the majority voting mechanism of multiple rounds of prediction results to achieve stable reinforcement learning under unsupervised conditions. The method is especially suitable for answer non-uniqueness scenarios such as text error correction (e.g., multiple reasonable solutions exist for punctuation correction), breaks through the limitation of the traditional single-solution model for math/code-based problems, and opens up a new path for test-time training of open-domain tasks. In the face of increasing unlabeled data and task diversity, exploring the adaptive learning ability of models under unambiguous supervision has become a key topic in the field.

## III. our model CEC-Zero:

### 3.1 Problem definition:

Chinese Spell Check (CSC) aims to correct all misspelled characters in the source sentence. Given that the source sentence $X = \{x_1, x_2,..., x_n\}$ contains n potentially misspelled characters, the model attempts to generate a target sentence of the same length $Y = \{y_1, y_2,..., y_m\}$ and correct all potential errors. The above process can be formulated as a conditional probability $P(Y|X)$. Specifically, for $x_i$, assuming it is an error character and its ground truth is $y_j$, the probability of correcting $x_i$ to $y_j$ can be written as, i.e., $P(y_j|X)$.

In the usual case, X and Y are of equal length, when for $x_i$, assuming it is an error character and its ground truth is $y_i$, the probability of correcting $x_i$ to $y_i$ can be written as, i.e., $P(y_i|X)$. In this type of case, since we have a stringent cleaning process and screening of the dataset, the idea of sequence labeling can be used to construct the loss function.

However, the actual Chinese spell checking is more complicated, and the sequence labeling approach cannot effectively determine and learn from unequal length sentences. As shown in the figure, we combine the source character $X = \{x_1, x_2,..., x_n\}$ and target sentences of different lengths $Y = \{y_1, y_2,..., y_m\}$ into a training data pair, at which point the BERT class-based model will not be able to effectively build a modeling framework, and at the same time, evaluating the model accuracy becomes complicated to the extent that we have to perform the necessary manual annotation.

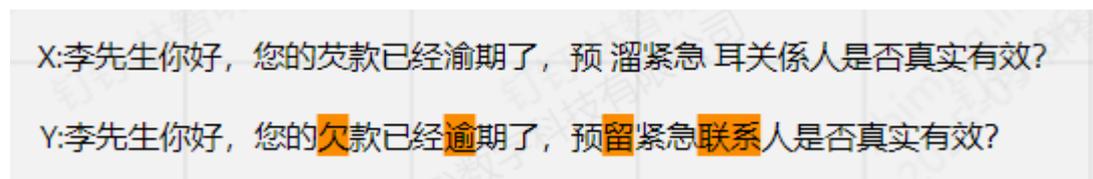

(A case of error correction in Chinese with unequal lengths of X and Y)

In order to solve the Chinese error correction problem from the source, especially the complex and intentionally diversified spelling errors made by human beings with practical applications, we have combed through most of the error paradigms visible in the existing datasets, and are quite ambitious to solve these problems at once. To this end, we came up with the idea of RL, which utilizes an innovative approach to creatively construct all the known misspelling paradigms to obtain the X required for the model, which together with the original sentence Y form the training data needed for LLM. This is because the cost of obtaining the original sentence Y in the real world is low and the amount of data is sufficient. This allows our method to work efficiently.

### 3.2 Theoretical approach:

Introduce the process of self-generated data: take the real-world original sentence as Y, and use some open-source and self-created text perturbation tools to get a number of X. The perturbation involves cases of homophone replacement, heteromorphic character replacement, merging of simple-structured Chinese characters, splitting of complex-structured Chinese characters, insertion of extraneous symbols, substitution of special symbols, as well as other perturbation paradigms that don't correspond to real-world usage. Suppose we generate m perturbed statements [X1,X2,... ,Xm-1,Xm], then we can collect {(X1,Y),(X2,Y),... ,(Xm,Y)} a total of m training data pairs.

Reward Functions for RL: Unlike common math and code problems, textual scenarios are more accommodating to the authentication of ground truth, which is due to the specificity of linguistic characters. As shown in the figure, we take the source character X = {x1, x2,... , xn}, the model generates two different target sentences Y1,Y2. Although there is only one comma sign difference between them, they are qualified error correction results in terms of semantics and practical applications.

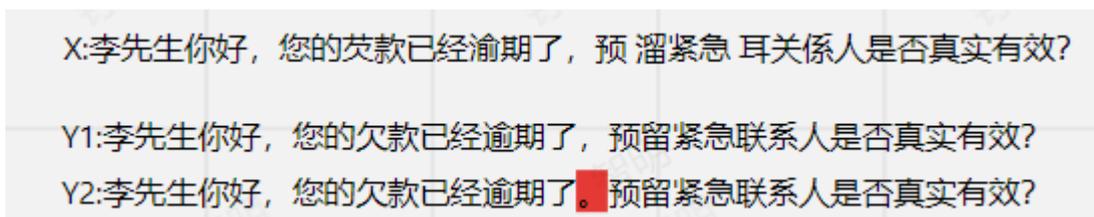

(Chinese error correction cases where Y1 and Y2 are different but both valid)

To address this challenge, we use sentence embedding to characterize the model's output Ypred and the real Y. The cosine similarity is used to evaluate the difference between Ypred and Y. We set the reward score to 1 for sentence pairs with cosine similarity of 1, and 0 for sentence pairs with similarity below the threshold theta, where theta is optional as a hyperparameter (default 0.8), and universally, we use the formula RLscore1 = max(0,(cos1-theta)/(1-theta)) to accomplish the reward score to complete the calculation of the bonus score.

In order to highlight the stability of the model and mitigate the model bias due to multiple real labels, inspired by Test-Time Reinforcement Learning (TTRL), we introduce the same mechanism to generate a batch of answers, and obtain a pseudo-answer based on the proximity of the spatial distance of the vectors within the answer to the embedding representation. Specifically, for an input X, the model generates [Y1,Y2,.... .Yl] for a total of l outputs, we obtain the corresponding embedding set {e1,e2,.... ...el}, and use the introduced functional function for finding the nearest distribution to find a {ex1,ex2,... ,exi} set of vector space points (where 1<=x1,.... .xj<=l) that satisfy this quality of being closest to each other's Euclidean distances, and the number of vectors in the set is greater than ⅓ the batch size requirement. We write e_center = avg(ex1,ex2,.... ,exi) to denote the center point of the set, i.e., the vector space representation of the pseudo-label. Following the previous approach, the cosine similarity is used to evaluate the e_center distribution with respect to e1,e2,.... . el to compute the cosine similarity. We set the reward score to 1 for sentence pairs with cosine similarity of 1 and 0 for sentence pairs with similarity below the threshold beta, where theta is optional as a hyperparameter (default 0.85), and universally, we use the formula RLscore2 = max(0,(cos2-beta)/(1-beta)) to accomplish the reward score calculation.

For the final reward function, we use RLscore = alpha*RLscore1 + gamma*RLscore2 to synthesize the final reward scores of the model. alpha and gamma are adjustable as hyper-references, with default values of 0.5, 0.5, respectively.

3.2 The idea of clustered scoring:

Unlike deterministic problems such as math and code, text-generated evaluations possess more complex logic and cannot use plurality and voting methods to make valid decisions. The reason for this is manifold: 1. An input X may get multiple acceptable Y's, which can be regarded as truthful labels even though they are different from each other; 2. Different truthful labels are difficult to be evaluated and scored using a simple symbolic judgment, because there may be only semantic similarity rather than strict symbolic consistency between different reasonable answers, and traditional symbol matching methods are difficult to be used directly for evaluation and quantitative scoring, while embedding technology provides a feasible basis for such evaluation by capturing the correlation at the semantic level.

We propose the idea of generating answer similarity clustering scores based on LLM - we intuit that convergent answers generated by human models are more likely to be accurate. We propose to use vector embedding clustering of sentences to determine the so-called "majority" answer, which is based on the premise that our experiments have confirmed that the model can get an accurate answer most of the time (i.e., most of the time it will find it accurate) and only a few times it will Only on a few occasions is the model wrong (i.e., a few outliers are incorrectly identified).

3.3 Description of the whole process:

Pseudocode generated from training data:

```
====================================
input (Y): true and correct sentences

output(X): textual scrambling, sentences with content to be corrected

1: Randomly select Y from a huge amount of datadata

2: Define the set of perturbation tools funtions

3: for Y in data:

4:      for func in functions:

5:          X = func(Y)

6:          train_data.append((X,Y))

7: return train_data
```

======================================

Training pipline flowchart!

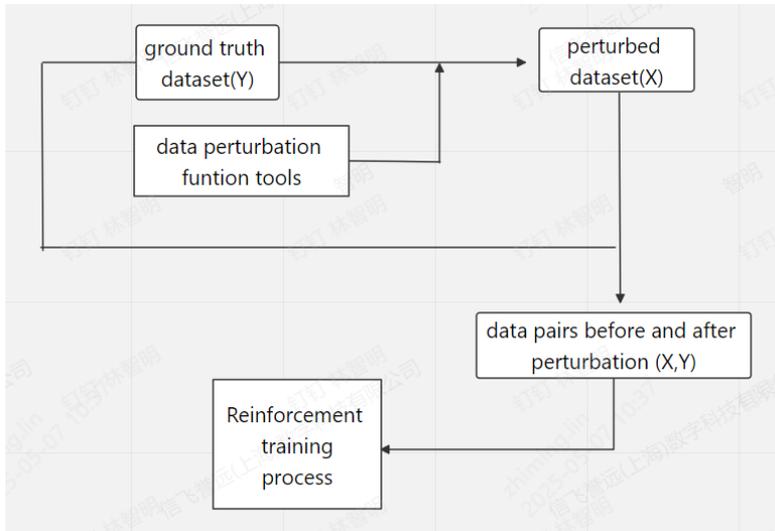

Reinforce the training process of learning:

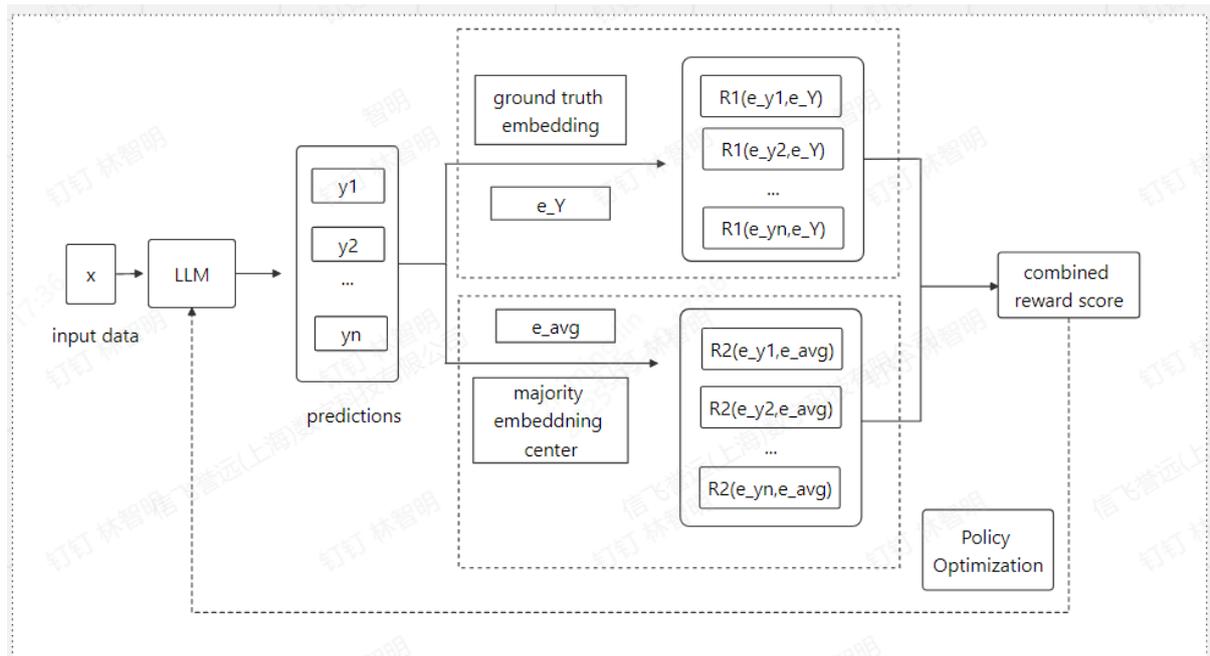

IV. Experiments:

4.1 Datasets and indicators.

Previous studies have chosen SIGHAN **[52]** as a benchmark. However, more and more studies, found that this dataset has many problems, such as semantic incoherence and annotation errors. Therefore, we choose two new CSC benchmarks, CSCD-NS **[64]** and LEMON **[17][57]** , in our study.(1) CSCD-NS: The quality of CSCD-NS is better than SIGHAN, and it is the first CSC dataset, in which the main source of character errors comes from Pinyin Input Method, which contains a large number of homophonic errors and word-level errors. (2) LEMON: LEMON is a novel, large-scale, multi-domain CSC dataset featuring a variety of real-world spelling errors. The dataset spans seven different sub-domains including games (GAM), encyclopedias (ENC), contracts (COT), medical (MEC), automotive (CAR), novels (NOV), and news (NEW), and is typically tested for the model's domain error-correction capabilities in a zeroshot setting.

We combined the training data from CSCD-NS and some of the data we collected in the customer service scenario as our training set. In addition, we collected web data from various domains on the web and cleaned it as additional training data. We use the validation data from CSCD-NS as the validation set and test the model on CSCD-NS test data, LEMON and our own customer service (CS) dataset respectively.

Evaluation Metrics: We report sentence-level and character-level precision, recall, and F1 scores to evaluate different models. These metrics are reported separately for detection and correction tasks. We compute the metrics using the CSCD-NS script. For predictions from the LLM that do not match the length of the source sentence, we first use ChERRANT to extract non-equal-length operations and then replace these operations with the source sentence before computing the metrics.

4.2 baselines.

We use the following CSC model for comparison.

The models in the BERT family are (1) BERT: BERT **[2][15][22][35]** treats CSC as a sequence annotation task **[7][12][19][45][71]** by encoding the input sentences and using a classifier to select appropriate characters from a vocabulary. (2) Soft Mask BERT (SMBERT) : SMBERT **[46][69]** consists of a detection and correction network that enhances the error detection capability of BERT. (3) SCOPE **[21]** : SCOPE combines an assisted pronunciation prediction task with an adaptive task weighting scheme to improve CSC performance.

For the selection of open source LLMs, we conducted a series of experiments using QWEN3 **[29][49]** . As one of the most powerful open-source LLMs, QWEN3 demonstrates strong Chinese processing capabilities and releases model parameters for multiple scales. We choose two models, Qwen3-14B (with thinking) and Qwen3-32B (with thinking), to evaluate the performance of generalized LLMs; for inference-based LLMs, we use Deepseek-R1-Distill-Qwen14B and Deepseek-R1-Distill-Qwen32B to evaluate the performance;

For closed-source LLMs, we evaluated model effects using ChatGPT , GPT-4 **[31]** , Doubao **[4]** , Claude 3.7, Gmini 2.5, and all LLMs participating in the test used the same cue word to perform the CSC task.

4.3 Main Results.

| Models | CAR | COT | ENC | GAM | MEC | NEW | NOV | CSCD-NS | CS | Avg |
|---|---|---|---|---|---|---|---|---|---|---|
| BERT | 46.87 | 52.61 | 45.74 | 23.41 | 42.73 | 46.63 | 32.25 | 65.49 | 27.94 | 42.63 |
| SMBERT | 49.91 | 54.85 | 49.33 | 26.18 | 46.91 | 49.16 | 34.56 | 67.22 | 44.67 | 46.98 |
| SCOPE | 50.71 | 54.89 | 45.23 | 24.74 | 48.72 | 48.72 | 33.17 | 71.7 | 43.82 | 46.86 |
| ChatGPT | 44.88 | 57.11 | 54.46 | 28.78 | 49.85 | 44.4 | 31.77 | 52.5 | 70.73 | 48.28 |
| GPT-4 | 54.44 | 62.82 | 55.12 | 36.27 | 56.36 | 56.09 | 45.64 | 54.41 | 80.48 | 55.74 |
| Doubao | 55.81 | 63.03 | 56.23 | 39.89 | 57.34 | 55.89 | 42.31 | 69.45 | 81.05 | 57.89 |
| Claude3.7 | 55.32 | 64.19 | 54.05 | 37.86 | 53.58 | 58.95 | 46.78 | 59.07 | 79.96 | 56.64 |
| Gmini2.5 | 56.01 | 61.27 | 55.8 | 40.12 | 54.89 | 61.04 | 41.97 | 66.29 | 81.04 | 57.60 |
| Qwen3-14B | 46.88 | 56.95 | 55.37 | 35.39 | 53.71 | 51.99 | 40.12 | 53.78 | 75.28 | 52.16 |
| Qwen3-32B | 52.97 | 57.45 | 55.41 | 36.03 | 53.23 | 59.81 | 44.25 | 58.97 | 81.09 | 55.47 |
| Deepseek-R1-Distill-Qwen14B | 53.07 | 56.85 | 55.89 | 38.95 | 55.19 | 53.04 | 43.1 | 60.18 | 79.86 | 55.13 |
| Deepseek-R1-Distill-Qwen32B | 55.57 | 63.52 | 55.03 | 39.29 | 56.63 | 55.93 | 44.77 | 67.32 | 85.39 | 58.16 |
| Qwen3-14B-RL(OURS) | 60.32 | 66.71 | 59.77 | 42.43 | 68.02 | 73.39 | 48.96 | 76.34 | 90.34 | 65.14 |
| Qwen3-32B-RL(OURS) | 63.28 | 66.89 | 61.30 | 44.29 | 74.87 | 79.91 | 51.29 | 79.71 | 91.78 | 68.15 |

V. Related work:

5.1 Test Time Scaling (TTS):

Test Time Scaling aims to improve the ability of large language models (LLMs) to handle complex tasks by increasing the computational resources available at test time. Previous studies have shown that TTS is more efficient than scaling during pre-training. Therefore, reallocating the same computational resources from pre-training to test time can lead to a greater improvement in model performance. Current research on TTS falls into two categories: parallel generation and sequential generation. Parallel generation involves the LLM generating multiple candidate responses (self-concordance, best-of-N, decision steps, or markers (reward-guided search)) during inference. Subsequently, an aggregation strategy is applied to integrate these candidates that typically use process reward models.

Meanwhile, sequential generation focuses on extending the output of LLMs to include longer responses with reflection and thought chain processes. While cueing techniques are widely used, they are often limited by the capabilities of the underlying model. Notably, DeepSeek-R1 is a representative advancement in this field, which enables extended reasoning capabilities in pre-trained language models through result-based reinforcement learning (RL), more specifically swarm relative policy optimization. In contrast to the first approach, which requires intensive process-level supervision, the second approach is more scalable due to its reliance on rule-based **[9]** rewards.

In addition to the above methods focusing on extending test time reasoning calculations, another method for increasing test time calculations is test time training (TTT). We will describe the relationship between these terms in Appendix A. While previous work has focused on applications such as video generation and comprehension, and to some extent on large-scale language models, the integration of test-time scaling with reinforcement learning remains largely underexplored.

5.2 RL for model inference:

The instruction-following capabilities of Large Language Models (LLMs), especially through methods such as Reinforcement Learning from Human Feedback (RLHF) **[25]** . RLHF uses algorithms such as Proximal Policy Optimization (PPO) **[30]** to match the underlying model to human preferences, where preference modeling is crucial. Recently, Large Reasoning Models (LRMs) such as DeepSeek-R1, exemplified by GRPO **[3]** , have demonstrated the relevance of RL **[50][67]** in improving reasoning using rule-based rewards.

RLHF is specifically designed for open-domain commands, whereas GRPO is specifically designed to inspire long chains of thought (CoT) in mathematical problem solving. Recent research has focused on improving the training stability of rule-based RL methods (such as PPO, GRPO and DPO **[42]**).

However, these methods typically train LLMs only on supervised training data, while inference involves generating extended CoT inference on unseen test problems. Furthermore, current RL methods rely on verifiable outputs - such as solutions in math or code - that can provide reliable reward signals. Such conditions are often impractical in real-world agent tasks, where verification is difficult, if not impossible (Wei et al., 2025a). Emerging research emphasizes a paradigm shift from learning solely from human labeled data to learning from empirical interactions. In this context, it becomes increasingly important for policy models to generate and label their own training data through real-world interactions.TTRL provides an initial attempt at RL with self-labeling incentives and moves toward learning from streams of experience.

5.3 Textual error correction:

Chinese text correction is a Chinese application **[12][14][16][19][21][33][34][41][45][47][57][61][63][70][72]** that is closely related to daily life. Due to the complex characteristics of the Chinese language, Chinese text error correction is a basic but challenging task. According to the different types of errors, Chinese text error correction is mainly categorized into two types, i.e., CGEC and CSC, and CGEC is further categorized into CSL CGEC and native CGEC according to the target user groups.There is a large gap between the language usage habits of foreigners and that of native Chinese speakers. This gap can lead to models trained or evaluated on these datasets that are not well adapted to a wider range of Chinese application scenarios. Therefore, researchers have recently begun to focus on CGEC tasks for native Chinese speakers . Compared with the relatively simple CSL CGEC task, the types of Chinese grammatical errors that the native Chinese CGEC task focuses on are more complex, such as structural confusion, illogicality, missing constituents, constituent redundancy, inappropriate sets, and inappropriate word order, etc. The CSC task mainly focuses on Chinese spelling errors due to the confusion of pronunciation and strokes.

In the field of grammatical error correction, in addition to Chinese English Grammar Error Correction (EGEC) has also received attention. Several studies have been conducted to evaluate the performance of ChatGPT on the EGEC task. Since the linguistic features of Chinese and English are inherently different, unlike previous work, our work focuses on Chinese, aiming to explore the Chinese error correction capability of LLM, and to promote the development and progress of Chinese text error correction in the LLM era .

BERT-style CSC models: With the advent of pre-trained language models, the dominant approach to CSC first shifted to BERT-style models, which treat CSC as a sequence annotation task. These models map each character in a sentence to its correct counterpart and fine-tune them to the sentence and counterpart pair. These models map each character in a sentence to its correct counterpart and fine-tune the sentence and the reference sentence. In addition, some studies have integrated phonological and morphological knowledge to improve the annotation process. However, due to parameter limitations, these models perform poorly in low-frequency and complex semantic scenarios compared to LLM.

Autoregressive CSC models can infer each marker in parallel, whereas autoregressive CSC models process markers sequentially. Previous research **[41]** suggests that autoregressive models such as GPT-2 **[1]** may not perform well on CSC. With the development of LLMs, several studies have investigated their text correction capabilities. These studies found that although ChatGPT knew the phonology of Chinese characters, they did not know how to pronounce them, which made speech error correction challenging. Speech error correction is challenging. Other studies have noted that ChatGPTs usually produce very smooth error correction but also bring more overcorrection. These findings are consistent with our observations and emphasize the need and importance of improving the performance of LLM on CSC.

VI. Analysis and discussion:

6.1 A fragile, overfitting, poorly generalized Chinese text error correction model does not bring practical application value, which is the point of concern in this article. Compared to the design of the results in the model or a small correction of some optimization method. We aim to find a more concise and generalized way to design our model. We focus on such 3 problems at the beginning of our design: 1) Textual Error Correction Because of the widespread existence of symbols and the existence of synonyms of words, often an incorrect sentence can correspond to multiple correct expressions, so how to design data and modeling for such scenarios is a major challenge.

2) The evaluation of utterance error correction is actually a very complex process, and there needs to be a balance between efficiency and accuracy to find a stable and efficient evaluation method. Theoretically, using manual annotation is a perfect solution, but it is too idealized and costly. And while simple word-by-word comparisons possess good engineering efficiency, they fail to address issues such as synonyms, symbols, etc., and are not accurate enough.

3) Existing datasets tend to be more fragile and the training data is cleaned too much, which is a gap with our original intention - to design a modeling service that has the value of being applied in industry, we need to address the dataset issues, which include diversity, complexity, data volume, etc.

6.2. The premise that RL can be widely used is that it can be effectively verified. In the field of Chinese error correction, it is costly to directly label the wrong sentences. Reasonable reverse thinking and error editing of correct sentences can enable us to obtain a large amount of high-quality training datasets in a short time. The diversity of editing methods can go as far as ensuring the information richness of the training dataset, which RL utilizes enough to form a generalization capability. The point that cannot be ignored here is that the edited sentences may have more than one ground truth. Therefore, we introduce the idea of clustering and scoring to generate a pseudo-label as an overall characterization of the potential set of ground truths, in order to avoid overfitting of the whole system to the trained dataset during the learning process. Although this problem can be solved with larger and more diverse training datasets, we believe that our approach is more cost-effective and efficient.

VII. Conclusion:

      This paper identifies a key shortcoming in current CSC learning, namely that traditional sequence labeling allows for over-conditionalized correction of errors, leading to limited generalizability. In this paper, we present CEC-Zero, a novel framework for training large language models using Reinforcement Learning (RL) on test data and propose a low-cost data generation methodology that eliminates the need for labor-intensive data collection and labeling.A key component of CEC-Zero is its answer clustering reward function. A component of CEC-Zero is its answer clustering reward function, which generates rule-based rewards based on the consensus between model predictions and the fact that there may only be semantic similarities between reasonable answers.

      Our experiments demonstrate the strong potential of CEC-Zero, with consistent improvements across a variety of models and tasks. We believe that CEC-Zero is a first step

towards the realization of RL with self-labeled rewards in non-standard scenarios such as text classes, marking an important direction in learning from a continuous stream of experience.